\documentclass[a4paper, 10pt, conference]{ieeeconf}      

\IEEEoverridecommandlockouts                              

\usepackage{graphicx} 
\usepackage{mathptmx} 
\usepackage{amsmath} 
\usepackage{amssymb}  
\usepackage{multirow} 
\usepackage{algorithm}
\usepackage[noend]{algpseudocode}
\usepackage{setspace}
\usepackage[table,usenames,dvipsnames]{xcolor}
\usepackage{overpic}
\usepackage{hyperref}

\usepackage{tikz,times}
\usetikzlibrary{positioning,decorations.pathreplacing,shapes,backgrounds}
\usetikzlibrary{shapes.geometric}
\usepackage{pgfplots}

\usepackage{microtype}

\usepackage{caption,subcaption}

\usepackage[english]{babel}
\usepackage{blindtext}
\usepackage{stfloats}

\title{\LARGE \bf
Learning Robotic Ultrasound Scanning Skills via Human Demonstrations and Guided Explorations
}

\author{Xutian Deng$^1$, Yiting Chen$^{1,2}$, Fei Chen$^{\dag2}$ and Miao Li$^{\dag1}$ 
\thanks{$^1$ School of Power and Mechanical Engineering, Wuhan University, Hubei, China (e-mail: \tt\small miao.li@whu.edu.cn)}
\thanks{$^2$ Department of Mechanical and Automation Engineering, T-Stone Robotics Institute, The Chinese University of Hong Kong, Hong Kong (e-mail: \tt\small fei.chen@iit.it)}
\thanks{$\dag$ Corresponding author}
}

\begin{document}

\maketitle
\thispagestyle{empty}
\pagestyle{empty}

\begin{abstract}
Medical ultrasound has become a routine examination approach nowadays and is widely adopted for different medical applications, so it is desired to have a robotic ultrasound system to perform the ultrasound scanning autonomously. However, the ultrasound scanning skill is considerably complex, which highly depends on the experience of the ultrasound physician. In this paper, we propose a learning-based approach to learn the robotic ultrasound scanning skills from human demonstrations. First, the robotic ultrasound scanning skill is encapsulated into a high-dimensional multi-modal model, which takes the ultrasound images, the pose/position of the probe and the contact force into account. Second, we leverage the power of imitation learning to train the multi-modal model with the training data collected from the demonstrations of experienced ultrasound physicians. Finally, a post-optimization procedure with guided explorations is proposed to further improve the performance of the learned model. Robotic experiments are conducted to validate the advantages of our proposed framework and the learned models.
\end{abstract}
\section{Introduction}
Medical ultrasound technology is widely adopted in clinical diagnosis due to its incomparable merits including non-invasive, low-hazard, real-time imaging,  safe and low cost. However, the performance of ultrasound examination is highly dependent on the skills and experience of sonographers, which generally requires a  large amount of time and effort to acquire \cite{arger2005teaching, hertzberg2000physician}. Moreover, the intensive and repetitive ultrasound scanning process causes a heavy burden on sonographers' physical condition \cite{murphy2000update}, further leads to the scarcity of ultrasound practitioners.

\begin{figure}[htbp]
\centering
\includegraphics[width=0.95\linewidth]{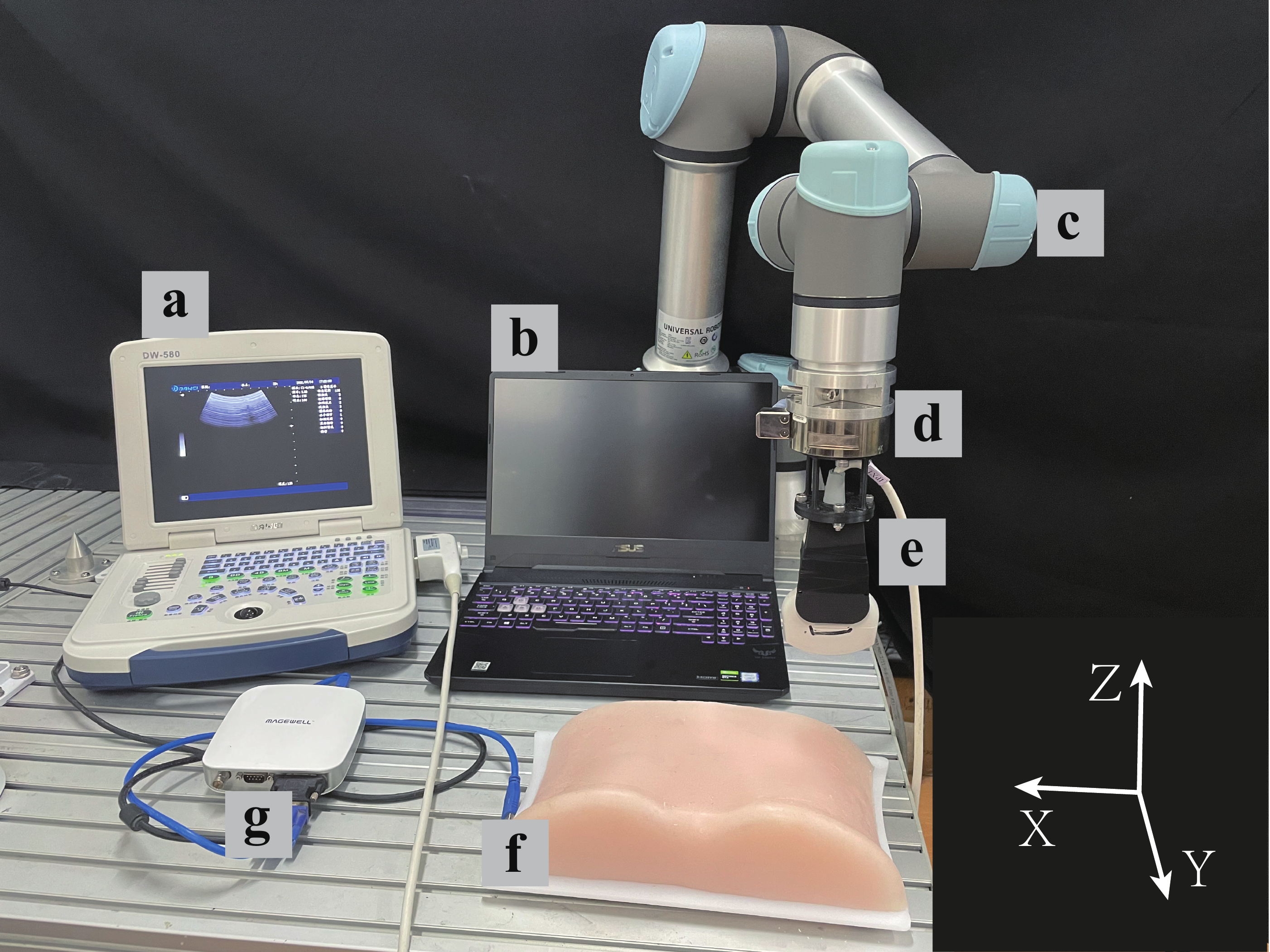}
\caption{\footnotesize The setup of our robotic ultrasound system. (a) The ultrasound machine. (b) The computer. (c) The collaborative robot UR5e. (d) The 6D force/torque sensor. (e) The ultrasound probe. (f) The kidney phantom. (g) The video capture device used to collect the ultrasound images.}
\label{fig::global_setup}
\vspace{-0.5cm}
\end{figure}

To address these issues, many research groups have attempt to use robotic ultrasound system (as shown in Figure.~\ref{fig::global_setup}) to help or even to replace the sonographers \cite{boctor2008three, priester2013robotic, chatelain20153d}. According to the autonomy level of robotic ultrasound system, they can be categorized into three types: tele-operated \cite{mathiassen2016ultrasound, guan2017study, sandoval2020cobot}, semi-autonomous \cite{mathur2019semi, victorova20193d} and full-autonomous \cite{virga2016automatic, kim2017development, huang2018robotic}. Obviously, a full autonomous ultrasound system is superior and our work in this paper follows this direction. However, it is still a challenging task to model robotic ultrasound scanning skills in a suitable manner so that the robot can execute it with similar dexterity as human sonographers. 

To this end, in this paper we propose to learn the ultrasound scanning skill from human demonstrations, which takes the ultrasound images, pose/position of the probe and contact force into account. The scanning skill is encapsulated into a multi-modal model and encoded as a deep neural network, which is trained with the demonstrated data. Furthermore, the obtained model is further improved with guided explorations to alleviate the negative effect from defective and biased demonstrations.

The main contributions of this paper are two-fold: (1)  an imitation learning-based multi-modal model encoded as a deep neural network is proposed to model the robotic ultrasound scanning skills, which takes the ultrasound images, contact force and the pose/position of the probe into account. (2) a novel post-optimization method with guided explorations is proposed to further improve the performance of the learned model in terms of generalization. This paper is organized as follows: The rest of this section presents some related work in the field of robotic ultrasound. Section \ref{sec::statement} introduces the proposed learning-based approach and the two-step training procedure. Section \ref{sec::experiment} describes the detailed experiments in real environment, with a final conclusion and discussion.

\subsection{Related Work}
\label{sec::related}

\begin{figure}[ht]
\centering
\includegraphics[width=0.8\linewidth, trim=0 100 0 0, clip]{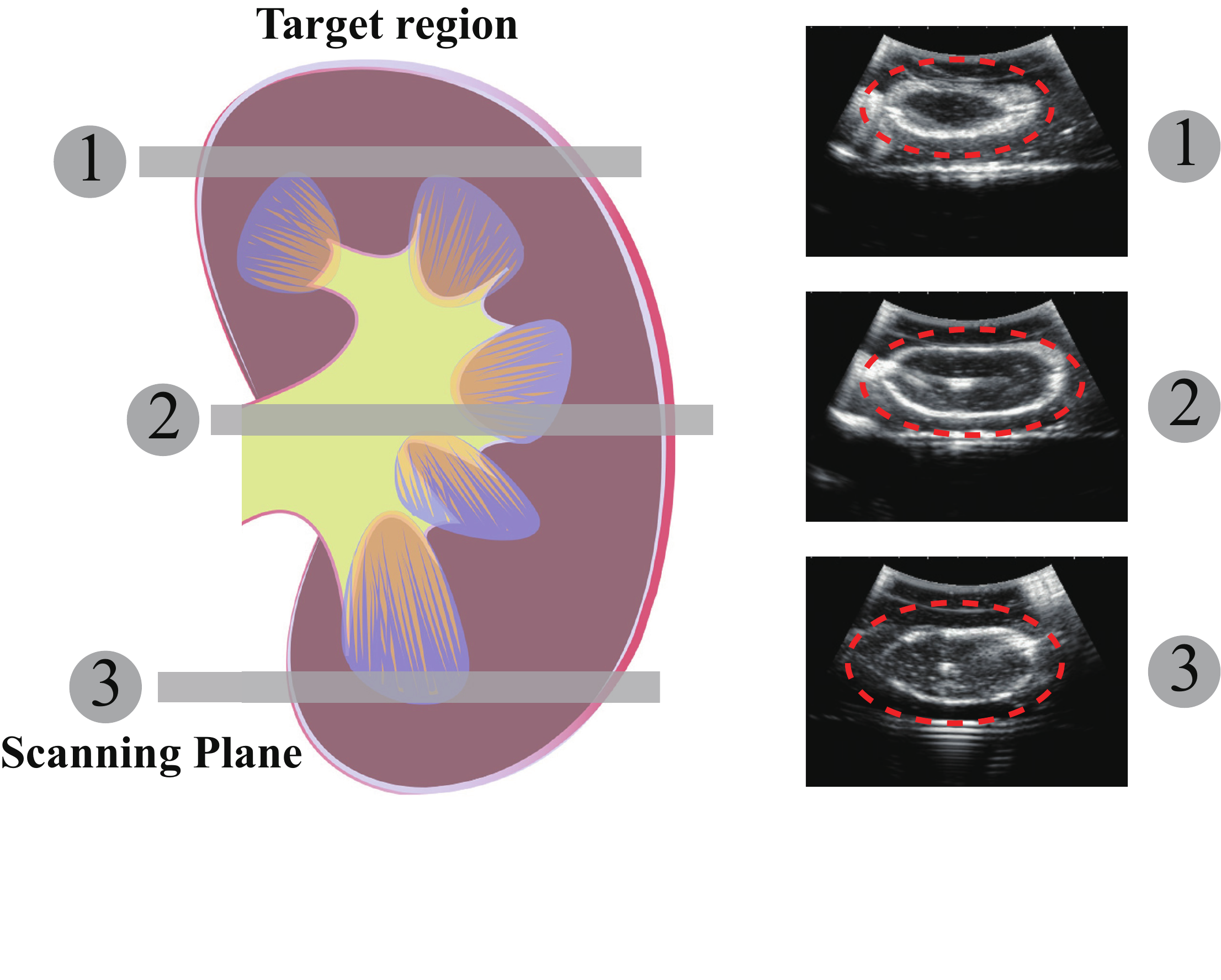}
\caption{\footnotesize Standard scanning planes of target region. The goal of our tasks is to autonomously acquire ultrasound images with the centered region of interest.}
\label{fig::task_description}
\vspace{-0.5cm}
\end{figure}

A variety of sensory feedbacks are involved during an ultrasound scanning process. Previous works have focused on integrating several sensor modalities together, with the goal to guide the scanning motion of the ultrasound probe.  For example, in \cite{chatelain2015optimization} a quality function is defined from the ultrasound image, which is used to control the motion of the probe. Different control policies based on feedback from ultrasound images have been proposed in \cite{hennersperger2016towards,patlan2017robotic,droste2020automatic,ning2021autonomic,li2021autonomous,milletari2019straight}. These methods either require a hand-coded function to describe the quality of the ultrasound image such as \cite{chatelain2015optimization}, or adopt a reinforcement learning-based approach that typically requires a large number of trials to obtain a specific scanning skill \cite{ning2021autonomic,li2021autonomous,milletari2019straight}. 

On the other hand, learning from demonstrations has been proved as an efficient approach to learn complex human skills \cite{schaal1999imitation}. In \cite{droste2020automatic}, an imitation learning-based approach is used to guide the probe movement for freehand obstetric ultrasound. In \cite{Huang2021}, the demonstrated ultrasound images are used to explicitly extract task features, which are further used during on-line execution with a visual servo controller. However, the contact force between the probe and the human skin has not been modeled in these works, which is known as an important factor for the quality of the final images \cite{arger2005teaching}. In the work of \cite{kim2017development,tirindelli2020force,jiang2020automatic}, force feedback is fused with ultrasound images for some specific purposes (i.e., localization of a vertebra \cite{tirindelli2020force}, finding normal direction of tissue surface \cite{jiang2020automatic}). In addition, for robotic manipulation, how to learn the compliant control policy for force-relevant tasks has been extensively studied in terms of force control and impedance control \cite{gao2019learning,li2018learning,li2014learning,zeng2020simultaneously}. A multi-modal representation for contact-rich tasks is propose in \cite{lee2019making}, consisting of the sensory information from vision and touch, which inspired our proposed framework significantly. To the best of our knowledge, this is the first unified framework that learns the robotic ultrasound scanning skills representation and the corresponding manipulation skills from human demonstrations including the modalities of ultrasound image, pose/position of the probe and the contact force.
\section{Problem Statement and Method}
\label{sec::statement}
\subsection{Problem Formulation of Ultrasound Scanning Tasks}

For each ultrasound scanning task, the key point is to find standard scanning planes of target region, as shown in Figure~\ref{fig::task_description}. The corresponding ultrasound images of standard scanning planes are with the centered region of interest, which means the most effective information for diagnosis is contained. Therefore, the focus of our research is to autonomously guide a robotic ultrasound system to acquire ultrasound images of standard scanning planes with the demonstrated ultrasound scanning skills. 

\begin{figure}[ht]
\centering
\includegraphics[width=1\linewidth]{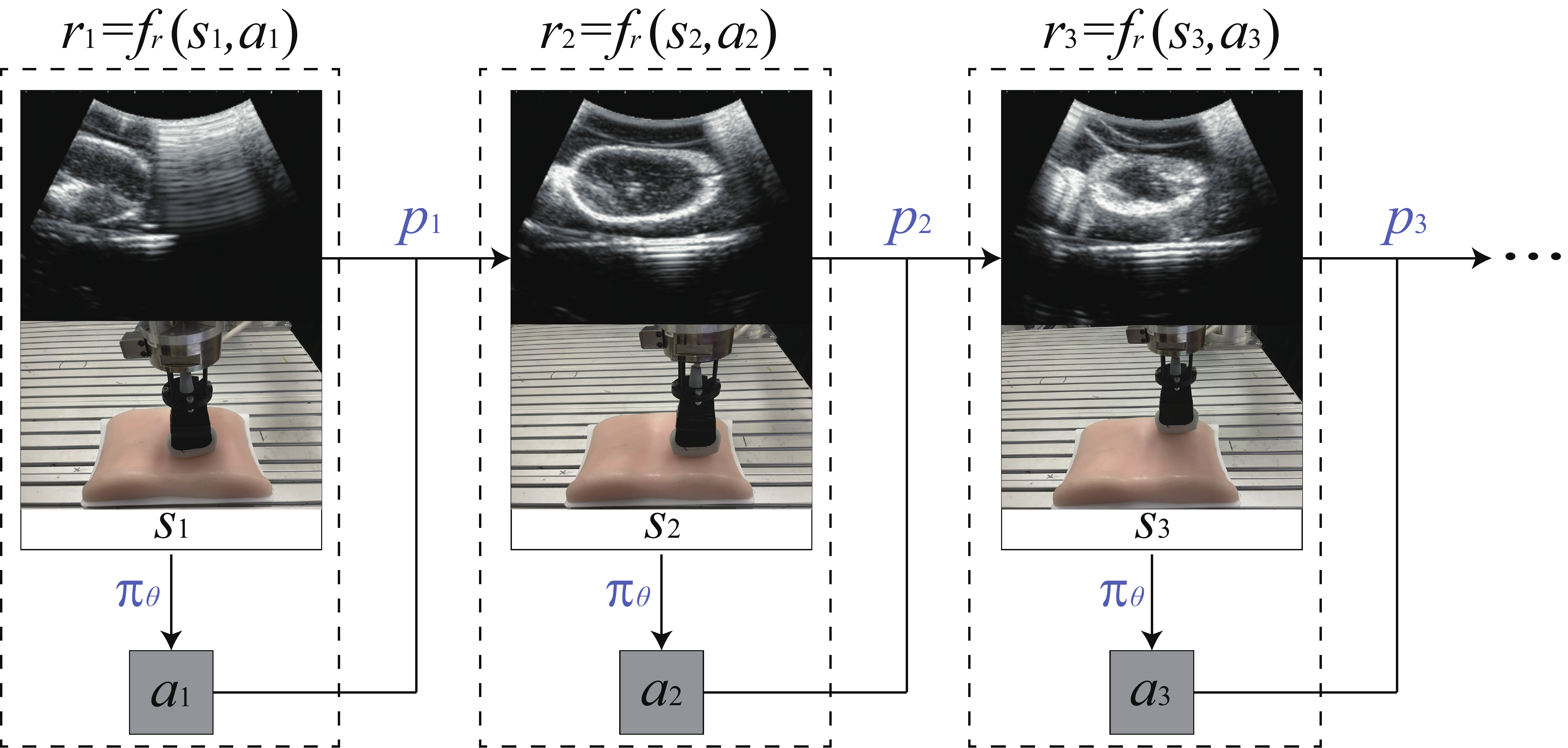}
\caption{\footnotesize The pipeline of MDP. According to the input state (ultrasound images, position of probe, pose of probe, and contact force), the policy should yield a befitting action.}
\label{fig::mdp}
\vspace{-0.5cm}
\end{figure}

As mentioned above, four different sensory modalities are closely related to the robotic ultrasound scanning skills, including: ultrasound images, position of probe, pose of probe, and contact force between the probe and the skin, which are described as follows:
\begin{itemize}
\item $D=\{(X_i,P_i,O_i,F_i)\}_{i=1...N}$ denotes a dataset with $N$ observations.
\item $X_i \in \mathbb{R}^{224\times224\times1}$ denotes the $i$-th collected ultrasound image with cropped size \footnote{$224\times224$ is the cropped image size for the ultrasound device in our work. For other types of ultrasound machine, different image sized can be chosen here.}.
\item $P_i \in \mathbb{R}^{3}$ denotes the position of the probe in the global coordinate system. 
\item $O_i \in \mathbb{R}^{4}$ denotes the pose of the probe in terms of quaternion. 
\item $F_i \in \mathbb{R}^{6}$ denotes the $i$-th contact force/torque between the probe and the human skin.
\end{itemize}

The ultrasound scanning process is modeled as a Markov Decision Process (MDP), which is described as follows:
\begin{itemize}
\item $M_i=<s_i, a_i, p_i, r_i>_{i=1...N}$ denotes the tuple about MDP.
\item $s_i \in S, s_i=<X_i, O_i, F_i>$ denotes the $i$-th state which is consisted of the ultrasound image $X_i$, probe pose $O_i$, and contact force $F_i$.
\item $a_i \in A, a_i=<\Delta P_i, \Delta O_i>$ denotes the $i$-th action which is consisted of position-difference $\Delta P_i$ ($\Delta P_i=P_{i+1}-P_i$) and pose-difference $\Delta O_i$ ($\Delta O_i=O_{i+1}-O_i$).
\item $p_i=f_p(s_i,a_i)$ denotes the transition function.
\item $r_i=f_r(s_i,a_i)$ denotes the reward function.
\end{itemize}
As shown in Figure~\ref{fig::mdp}, with the target to perform autonomous ultrasound scanning process, the ultrasound scanning skills is represented as a policy function $\pi (s)\to a$, which denotes the mapping from the current state $s$ to the predicted action $a$. The modeling and learning of policy $\pi$ about the ultrasound scanning skills is described in the following section.

\subsection{Ultrasound Skills Modeling and Learning}

\begin{figure*}[ht]
\centering
\begin{minipage}{0.35\linewidth}
\centering
\begin{tikzpicture}
\node at (-2,0) {\includegraphics[width=1\linewidth]{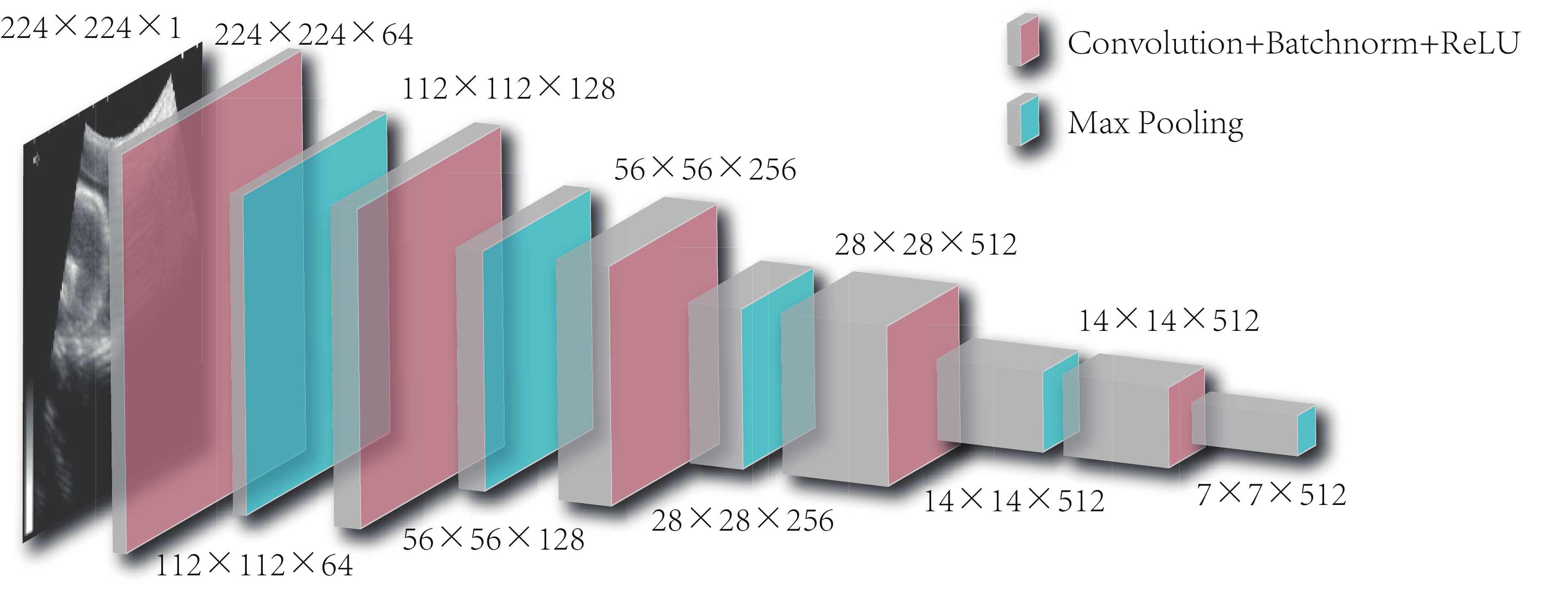}};
\node at (-2,1.5) {\tiny Convolution layers};
\draw[dashed] (-5.2, -1.2) rectangle  (1, 1.3);
\end{tikzpicture}
\end{minipage}
\begin{minipage}{0.6\linewidth}
\centering
\includegraphics[width=1\linewidth]{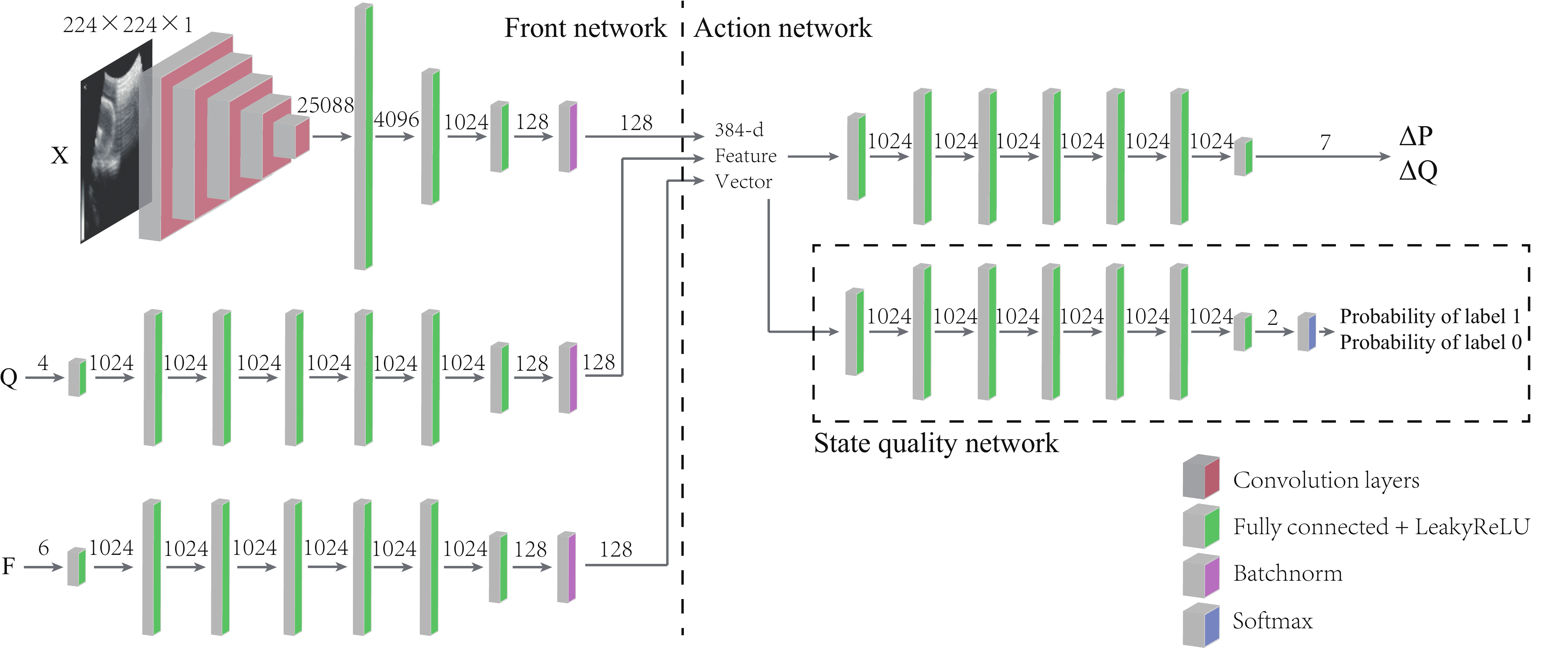}
\end{minipage}
\centering
\caption{\footnotesize Architecture of the neural network. The input information includes the ultrasound images $X$, the probe pose $Q$ and the contact force $F$. The output of the neural network is the predicted action, including the position movement $\Delta P$ and the pose movement $\Delta Q$. Further, we design a parallel network (the state quality network) to predict the probability of 1/0 label, and the confidence of state is defined as the probability of label 1.}
\label{fig::network}
\vspace{-0.5cm}
\end{figure*}

\begin{figure}[hb!]
\centering
\includegraphics[width=0.8\linewidth, trim=0 0 0 0, clip]{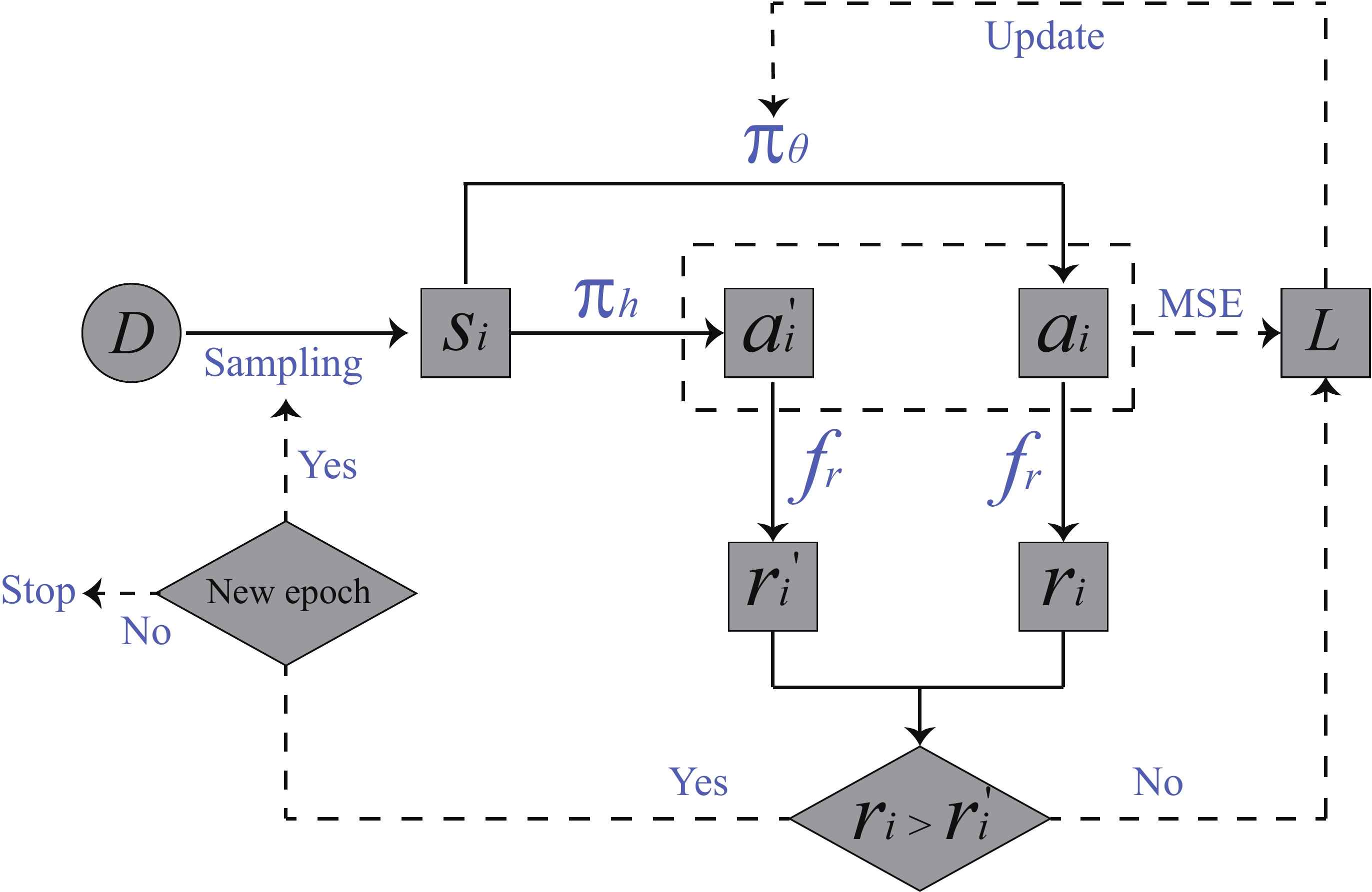}
\caption{\footnotesize The pipeline of the proposed post-optimization method. The optimized model will tend to generate higher-reward action for better task performance.}
\label{fig::pipeline}
\vspace{-0.5cm}
\end{figure}

We proposed to use a deep neural network as shown in Figure.~\ref{fig::network} to encode the policy $\pi$ with all the trainable parameters denoted as  $\theta$. The front network was designed to extract the ultrasound image $X$, the probe pose $Q$ and the contact force $F$ into three 128-d feature vectors. The probe position $P$ is hided to ensure that the model can extract position information from the ultrasound image $X$ and thereby strengthen its generalization. The action network catches the 384-d feature vector, generated by concatenating with three 128-d vectors, and yields the predicted position motion $\Delta P$ and pose motion $\Delta Q$. Besides, the state quality network is employed to identify the label of the current state, with acceptable/unacceptable states denoted as 1/0. The acceptable states contain ultrasound images with centered region of interest, while the unacceptable states are the opposite. In the following two subsections, a two-step training process is proposed to learn the policy $\pi_{\theta} (s)\to a$ from human demonstrated data.

\subsection{Learn the Skill Policy from Experience}
To train the model with some low-level exploration strategies, we proposed to leverage the power of imitation learning to pre-train the neural network. The training process of a model with parameters $\theta$ is described as follows:
\begin{equation}
{{\theta }_{t+1}}\leftarrow {{\theta }_{t}}-\alpha {{\nabla }_{\theta }}\sum\limits_{i}{L({{\pi}_{\theta}}({{s}_{i}}),{{a}_{i}}|{{s}_{i}}\in S,{{a}_{i}}\in A)}
\label{eqn_1}
\end{equation}
where $\alpha$ and $L$ denote the learning rate and the loss function respectly. Detaily, we choose the mean squared error as the loss function, which is described as follows:
\begin{equation}
L({{a}_{i}},{{a}_{j}})={{({{a}_{i}}-{{a}_{j}})}^{2}}
\label{eqn_loss}
\end{equation}
where $a_i$ and $a_j$ denotes two actions respectively.

The advantage of imitation learning is that the model can quickly learn the demonstrated tasks, as skipping the self-exploration stage. However, due to lack of reward function $f_r$ during training, the strategy of the trained model is too monotonous, with a large number of bias skills learned from the deficient demonstrations. To solve these problems, we propose a post-optimization method to train a skills-learned model, which is described in the following section.

\begin{figure}[b!]
\centering
\includegraphics[width=0.8\linewidth, trim=0 0 0 0, clip]{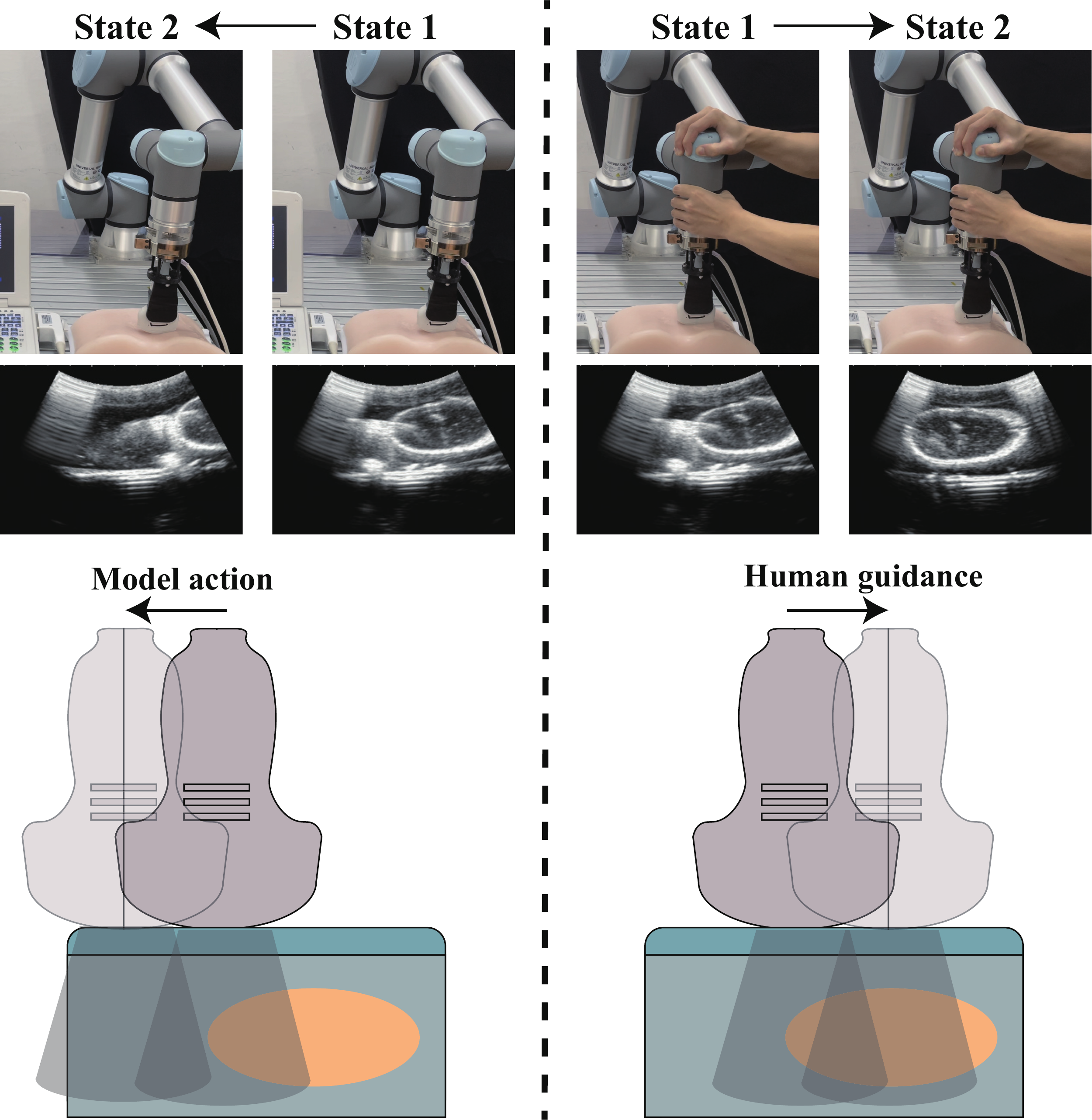}
\caption{\footnotesize The human guided explorations for post optimization. We propose to guide the pre-trained model with human guidance to improve the performance of the learned model.}
\label{fig::guided_exploration}
\vspace{-0.5cm}
\end{figure}

\subsection{The Post-optimization Method}
While the goal of pre-trained model is to clone behaviors of demonstrators, it has no guarantee that the learned skills can lead to an ultrasound scanning model with acceptable performance. To address this issue, a post-optimization method is proposed to improve the learned scanning skill, which is described as follows:
\begin{equation}
\footnotesize
{{\theta }_{t+1}}\leftarrow {{\theta }_{t}}-\alpha {{\nabla }_{\theta }}\sum\limits_{i}{L({{\pi }_{\theta }}({{s}_{i}}),\underset{a}{\mathop{\arg \max }}\,{{f}_{r}}({{s}_{i}},a)|{{s}_{i}}\in S,a\in \{{{\pi }_{\theta }}({{s}_{i}}),{{\pi }_{h}}({{s}_{i}})\})}
\label{eqn_2}
\end{equation}
where $\pi_h(s_i)$ denotes the human strategy. For the current state $s_i$, the pre-trained model could output the predicted action $a_i$ by $\pi _ \theta (s_i) \to a_i$. During the guided explorations, the strategy $\pi _ \theta$ will tend to learn and adopt a higher-reward action by comparing rewards of the predicted action $a_i$ and the human suggested action $a_{i}^{'}$. The reward function $f_r$ is described as follows:
\begin{equation}
{{f}_{r}}(s_i,a_i)=q(f_p(s_{i},a_{i}))-q(s_i)
\label{eqn_3}
\end{equation}
where $q\in [0, 1]$ denotes the state quality, which equals to the confidence of label 1 yielded from the state quality network presented in Figure~\ref{fig::network}. The detailed post-optimization method is presented in Figure~\ref{fig::pipeline}. For pre-trained model's low-reward actions, human guidance will help the model to move toward target region to generate higher-reward actions, thereby improve its strategy, as shown in Figure~\ref{fig::guided_exploration}.

\section{Experiments and Results}
\label{sec::experiment}

\subsection{Experimental Setup}

\begin{figure}[!t]
\centering
\includegraphics[width=0.9\linewidth]{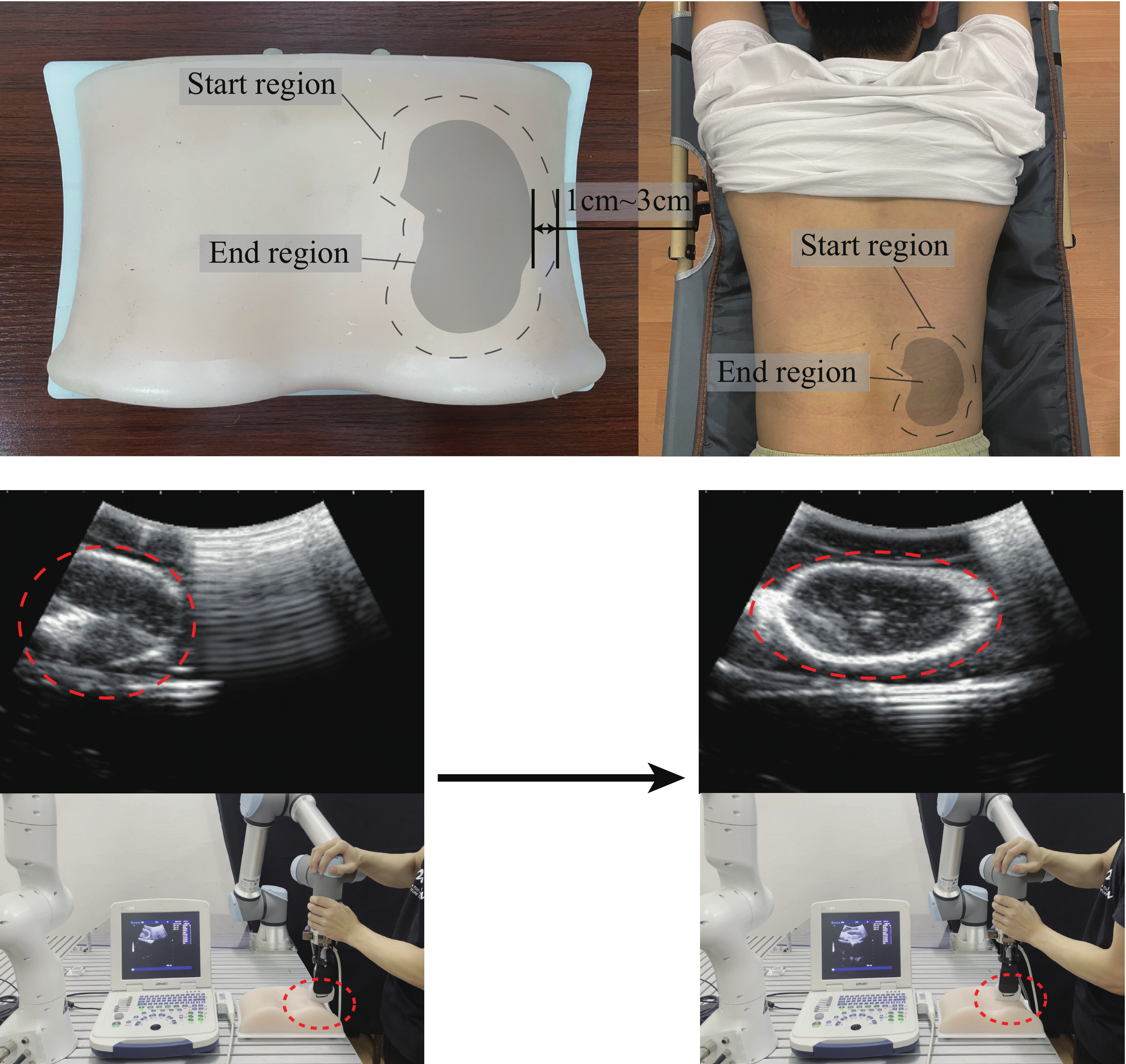}
\caption{\footnotesize The data collection of ultrasound scanning tasks. Demonstrations would finish when the ultrasound images of standard scanning planes were acquired.}
\label{fig::human_phantom}
\vspace{-0.5cm}
\end{figure}

\begin{figure*}[b]
\centering
\begin{minipage}{0.24\linewidth}
\centering
\includegraphics[width=1\linewidth, trim=40 0 80 0, clip]{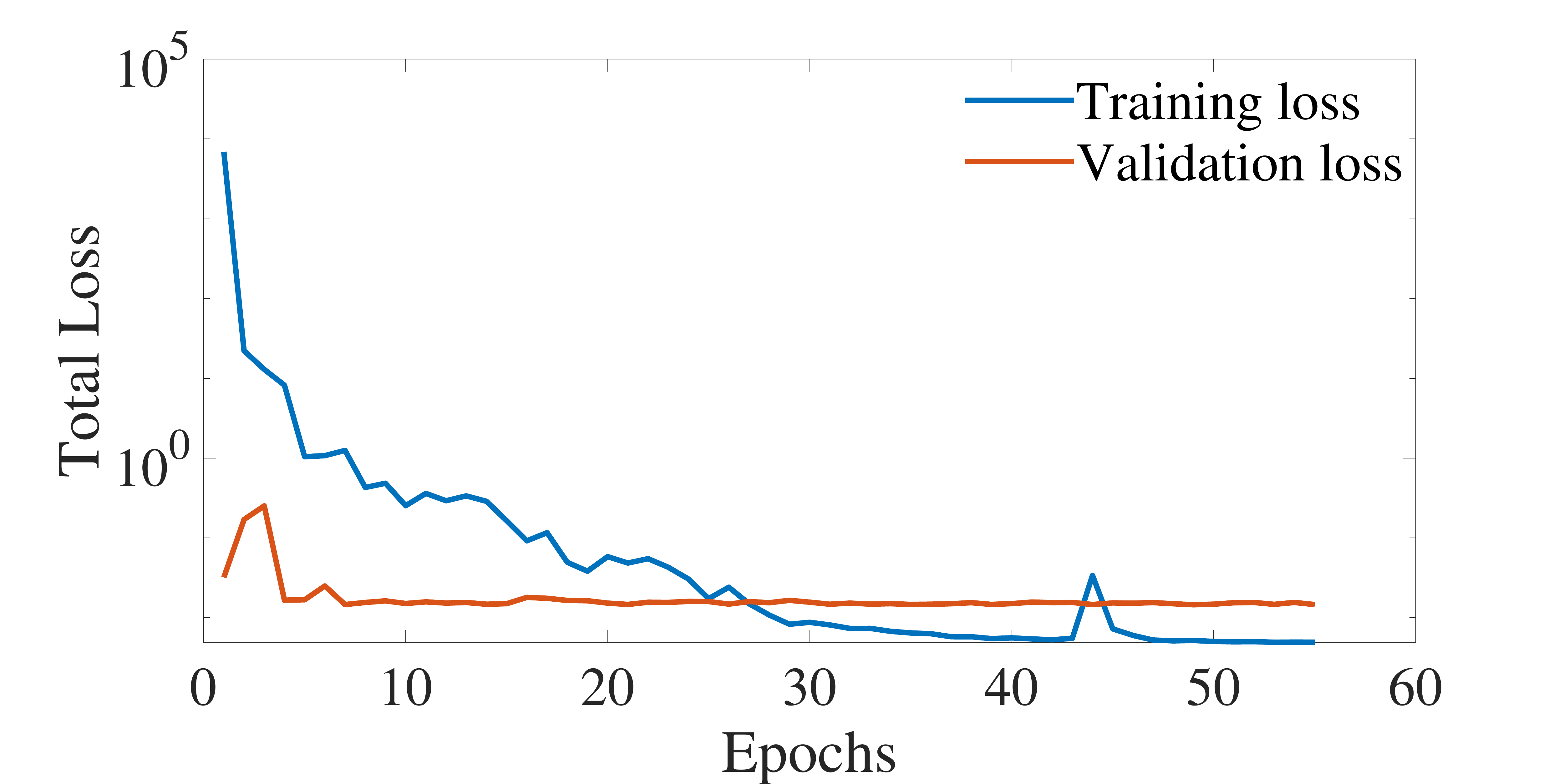}
\caption*{(a)}
\label{fig::pred_and_true_action::a}
\end{minipage}
\begin{minipage}{0.24\linewidth}
\centering
\includegraphics[width=1\linewidth, trim=80 0 80 0, clip]{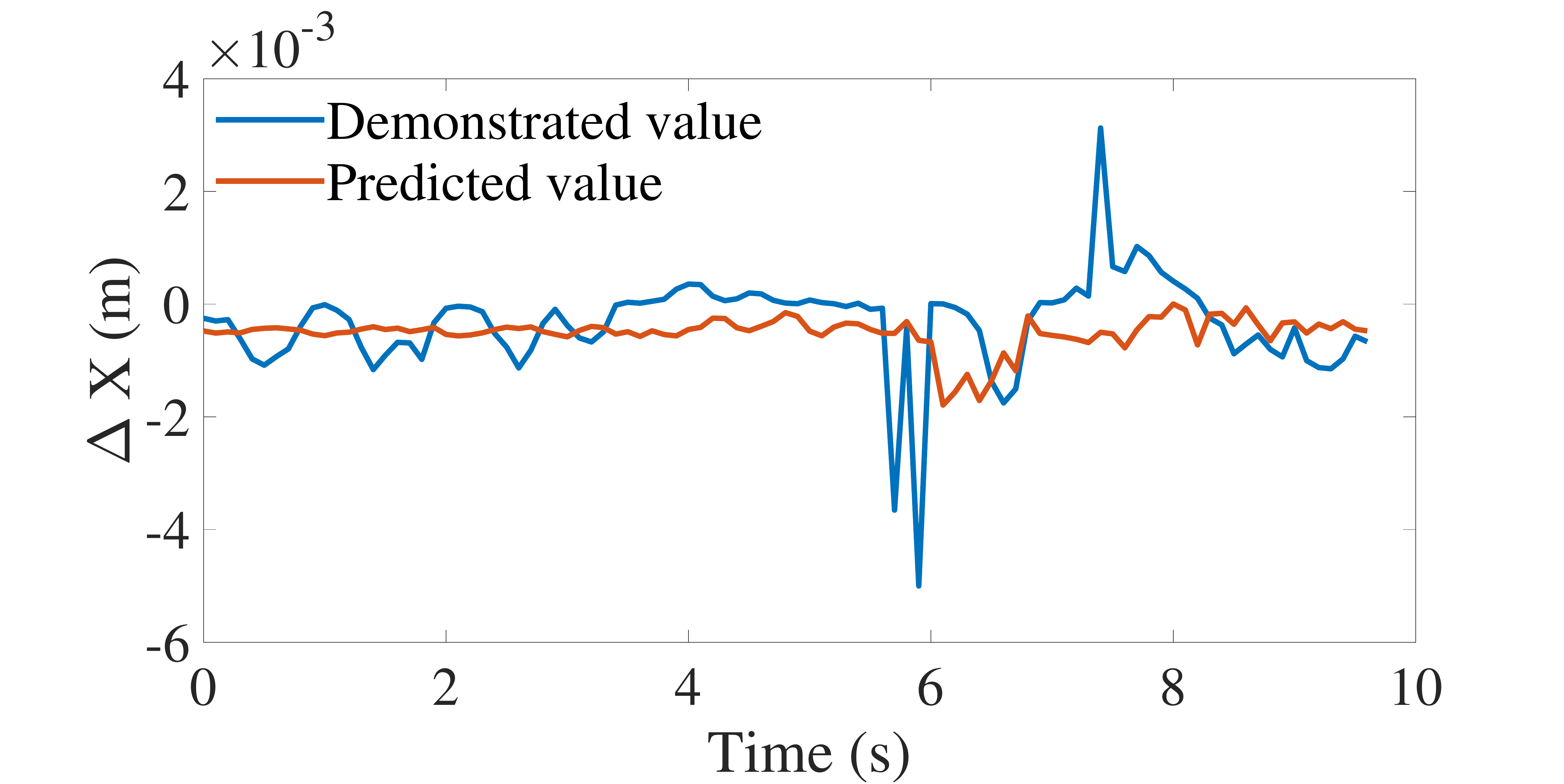}
\caption*{(b)}
\label{fig::pred_and_true_action::b}
\end{minipage}
\begin{minipage}{0.24\linewidth}
\centering
\includegraphics[width=1\linewidth, trim=80 0 80 0, clip]{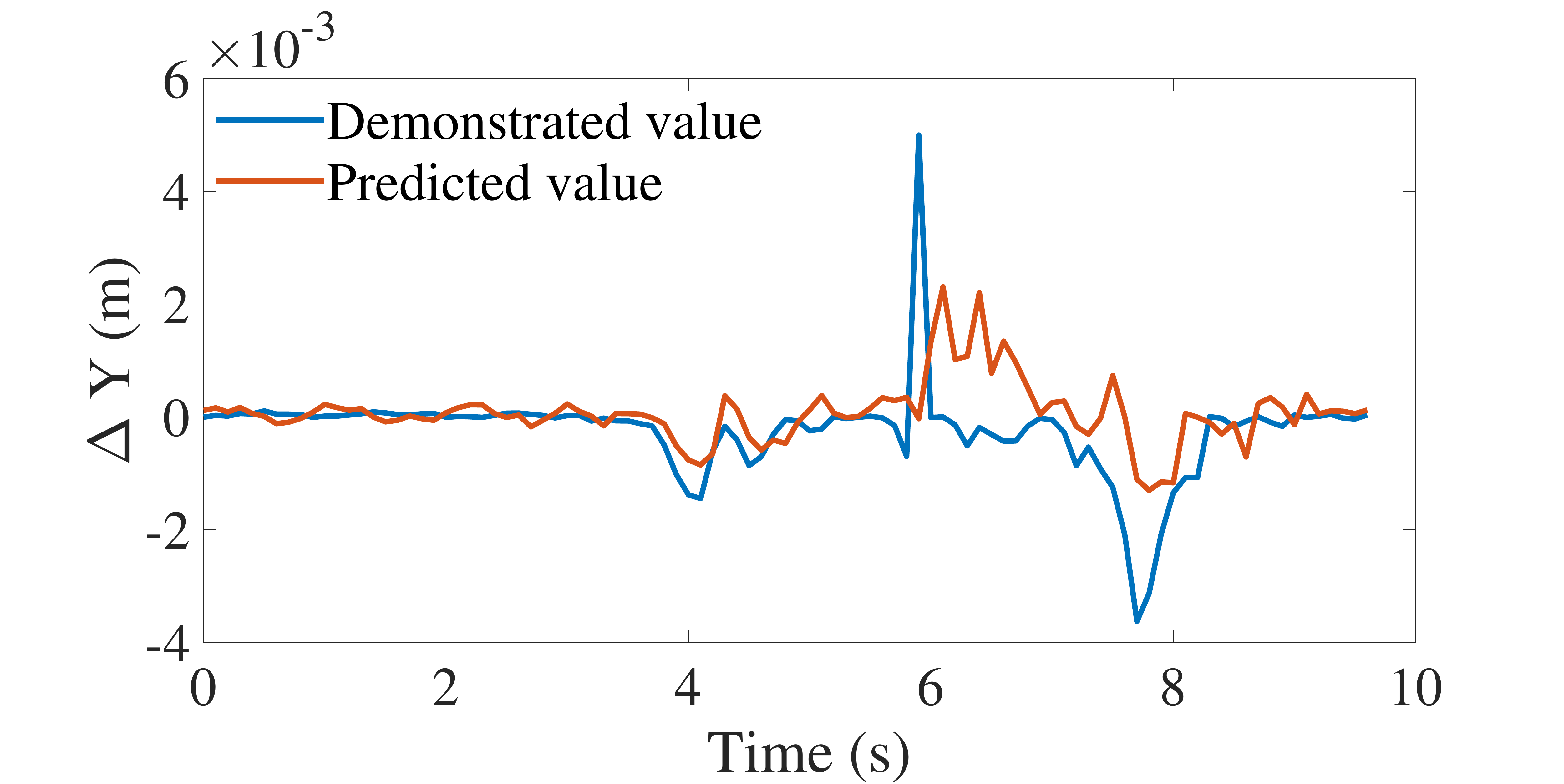}
\caption*{(c)}
\label{fig::pred_and_true_action::c}
\end{minipage}
\begin{minipage}{0.24\linewidth}
\centering
\includegraphics[width=1\linewidth, trim=80 0 80 0, clip]{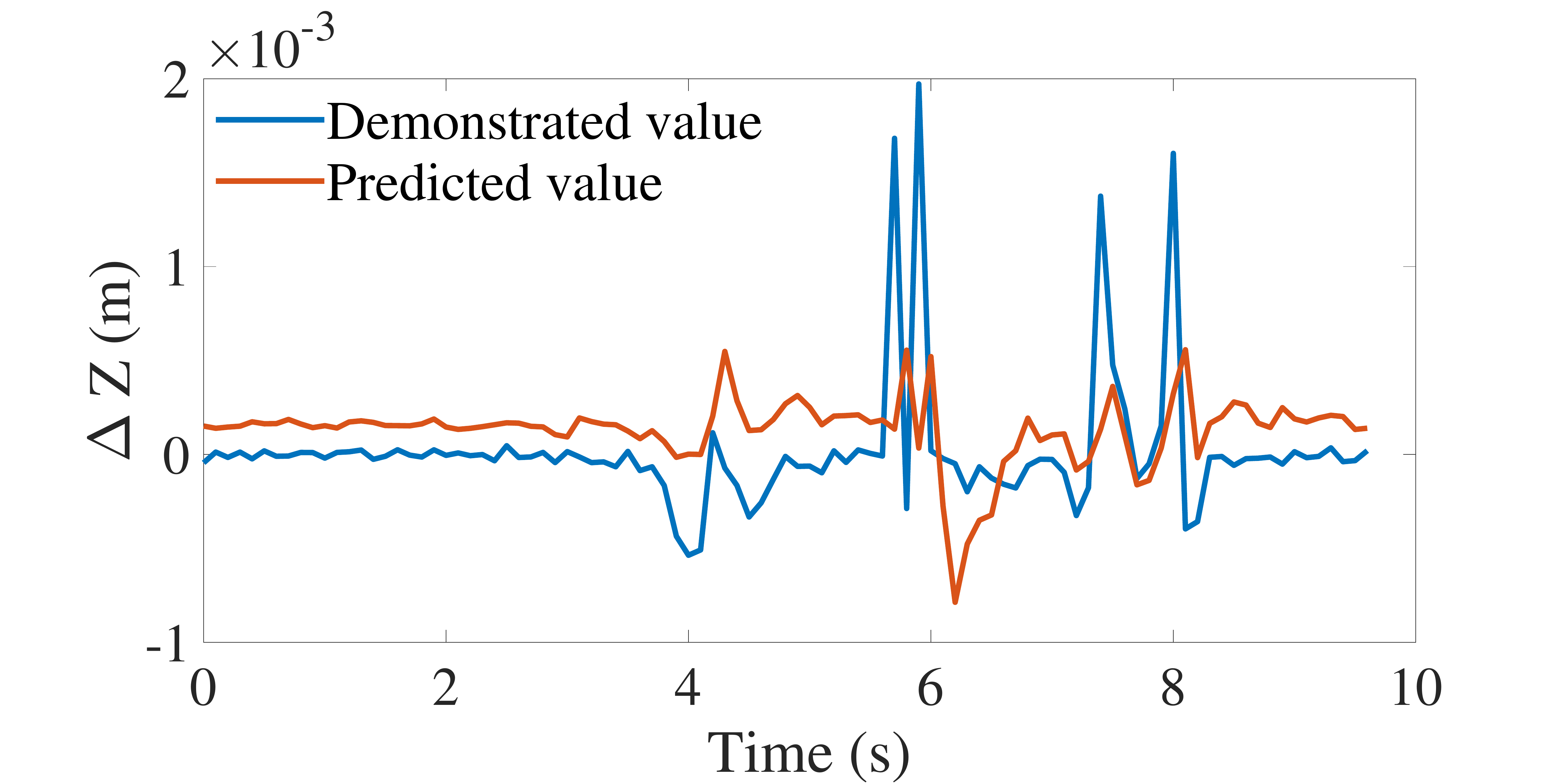}
\caption*{(d)}
\label{fig::pred_and_true_action::d}
\end{minipage}
\begin{minipage}{0.24\linewidth}
\centering
\includegraphics[width=1\linewidth, trim=50 0 100 0, clip]{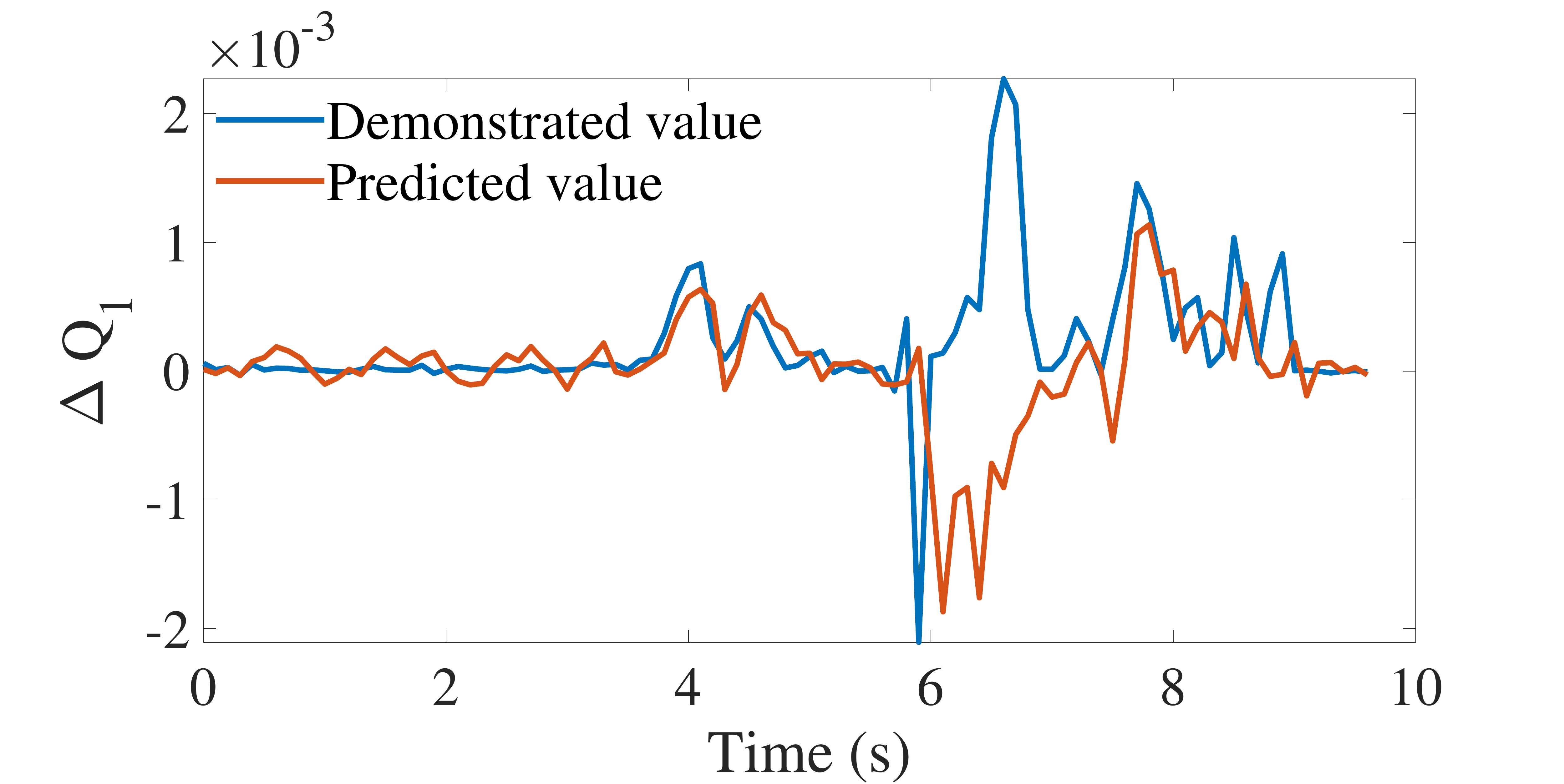}
\caption*{(e)}
\label{fig::pred_and_true_action::e}
\end{minipage}
\begin{minipage}{0.24\linewidth}
\centering
\includegraphics[width=1\linewidth, trim=50 0 100 0, clip]{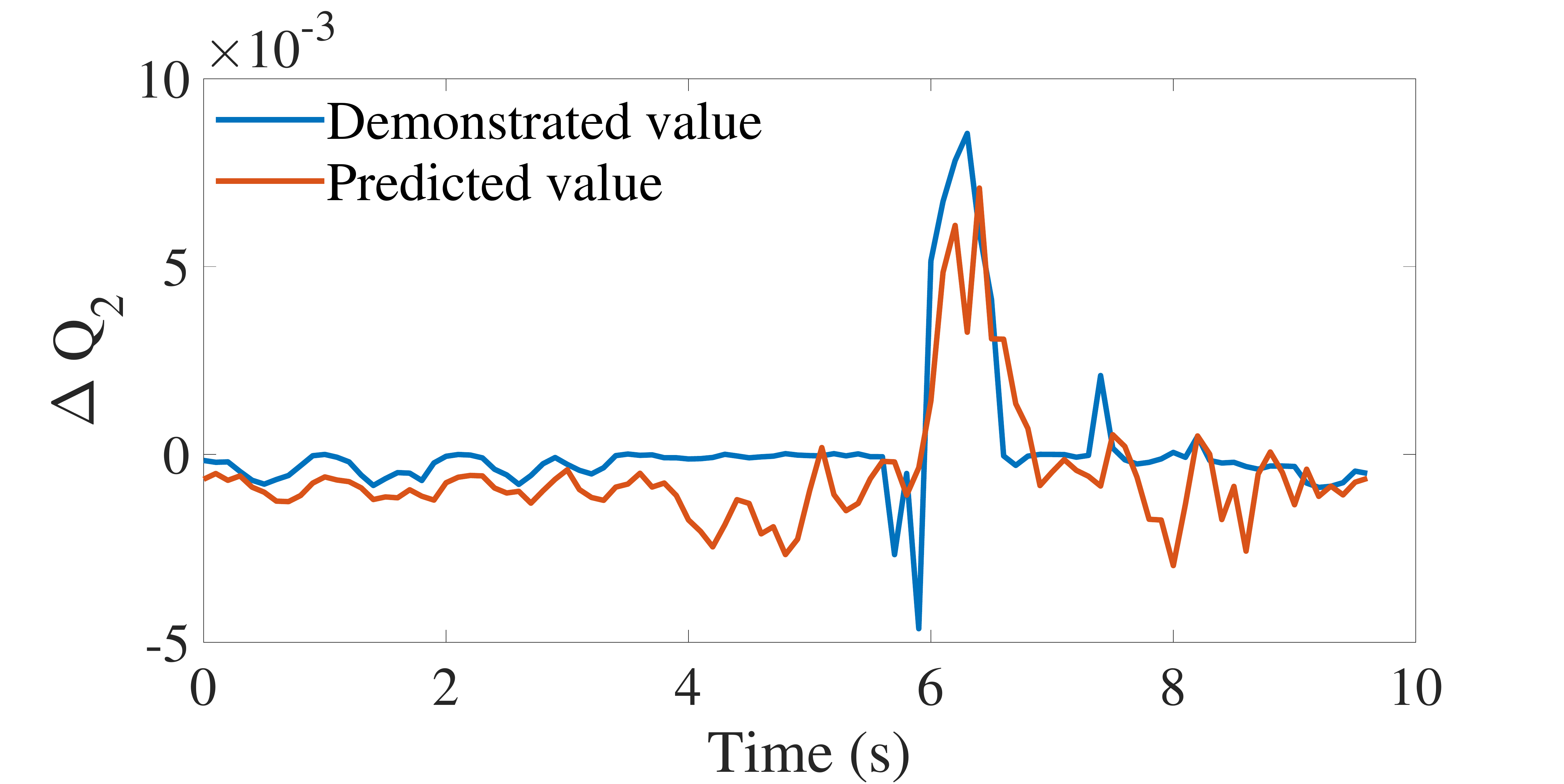}
\caption*{(f)}
\label{fig::pred_and_true_action::f}
\end{minipage}
\begin{minipage}{0.24\linewidth}
\centering
\includegraphics[width=1\linewidth, trim=50 0 100 0, clip]{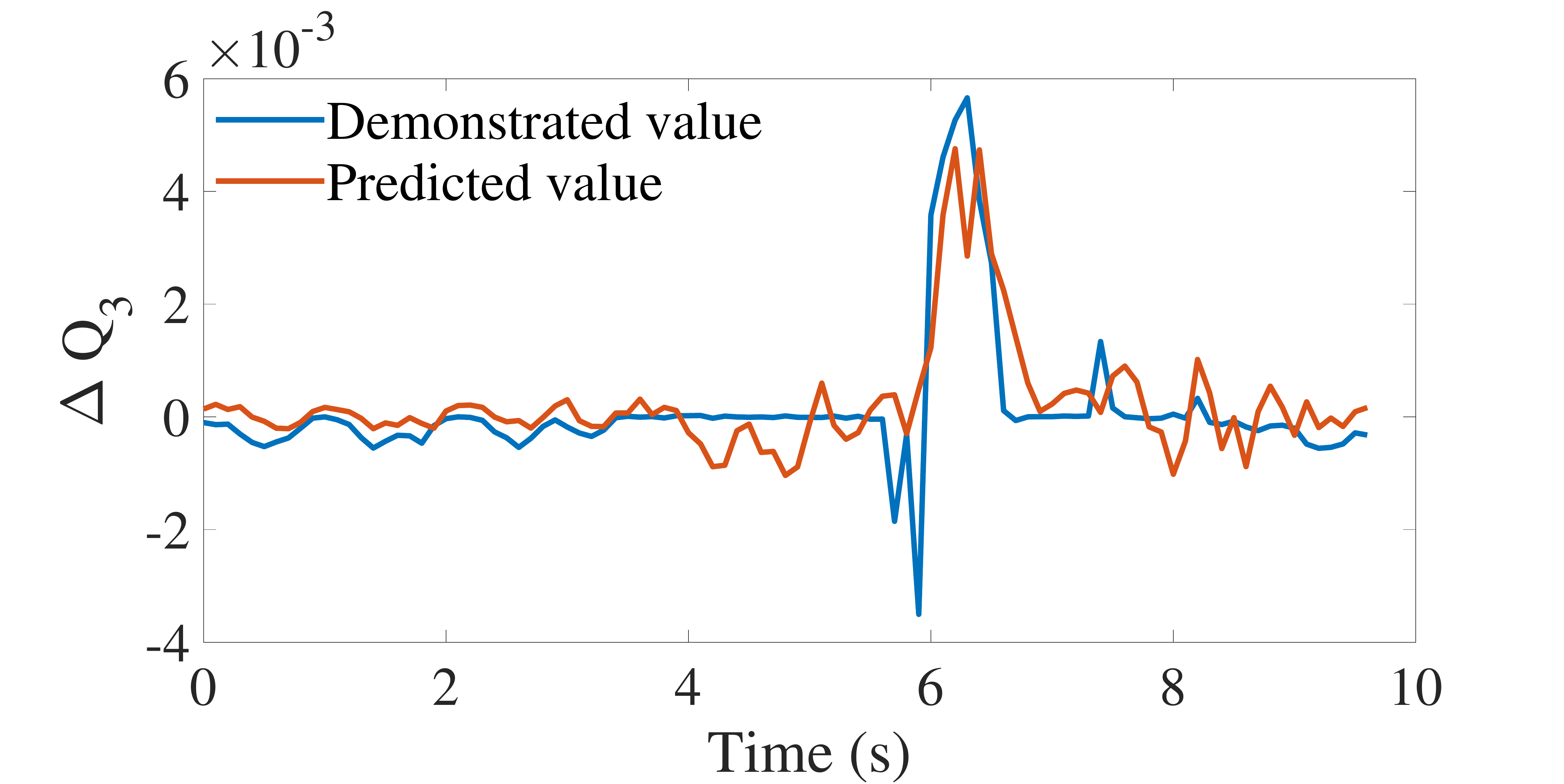}
\caption*{(g)}
\label{fig::pred_and_true_action::g}
\end{minipage}
\begin{minipage}{0.24\linewidth}
\centering
\includegraphics[width=1\linewidth, trim=50 0 100 0, clip]{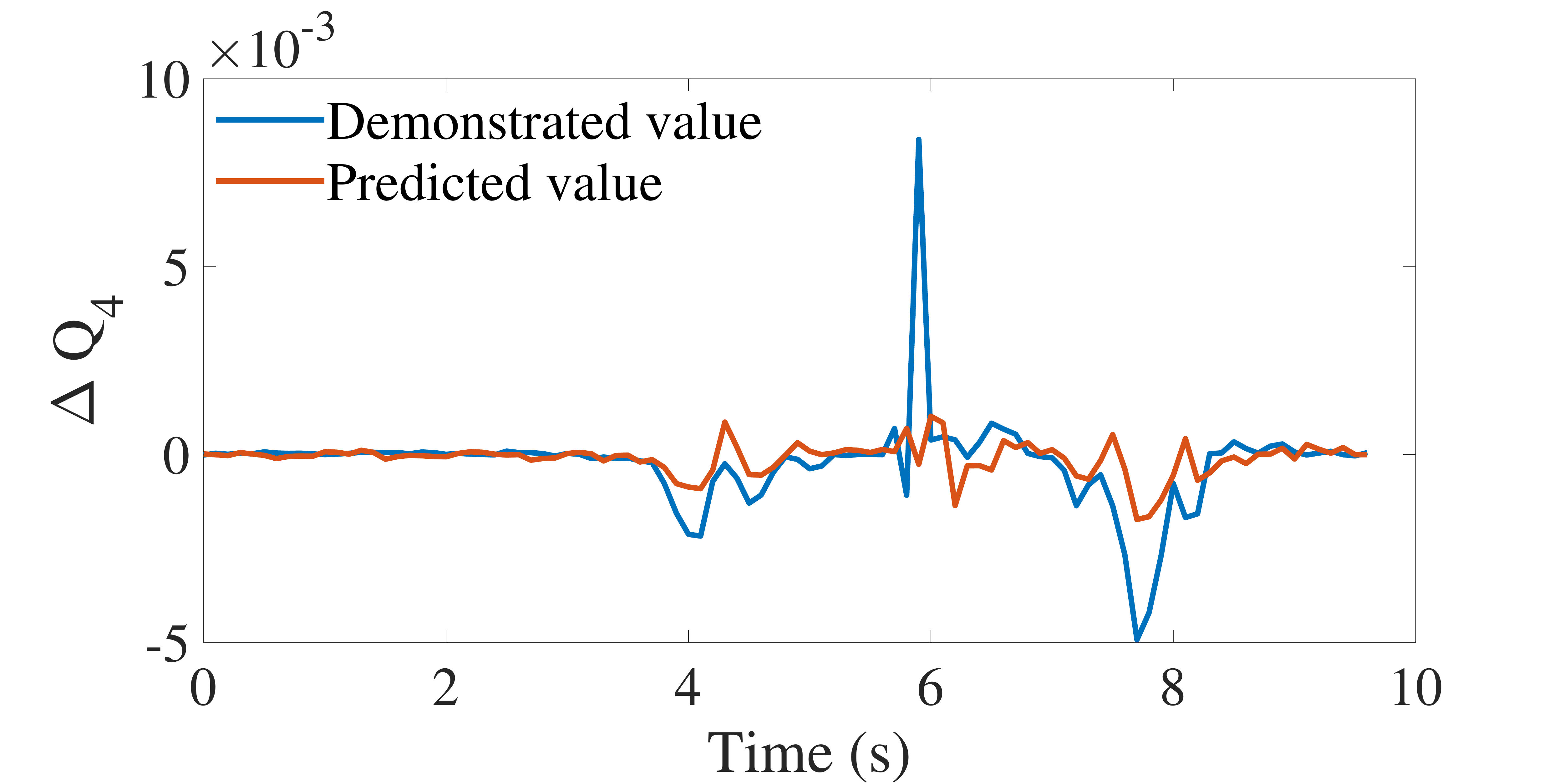}
\caption*{(h)}
\label{fig::pred_and_true_action::h}
\end{minipage}
\caption{\footnotesize The training and validation of network during imitation learning. (a) Loss of training and validation. (b-h) Demonstrated and predicted values of actions.}
\label{fig::pred_and_true_action}
\vspace{-0.5cm}
\end{figure*}

For ultrasound images, we used the MAGEWELL USB Capture AIO to capture videos from the DAWEI DW-580 ultrasound machine. The ultrasound probe was combined with the 6-degree force/torque sensor, which was mounted at the end of the collaborative robot UR5e. The ultrasound probe was controlled by the robot to scan the kidneys phantom. The sensory information was recorded and processed by a PC with Intel i5 CPU and Nvidia GTX 1650 GPU, with an environment of Ubuntu16.04 LTS and PyTorch 1.8.1. The data sampling rate is 10Hz and the robot control rate is 2Hz. The original image size is $640\times 480$, which is cropped as $224 \times 224$ for calculation.

\subsection{Data Collection}

During the data collection, the ultrasound probe was manually located in the start region of the phantom, as shown in Figure~\ref{fig::human_phantom}, which is an approximate annular area around the right kidney with a width of 1-3 cm. The scanning tasks were demonstrated by dragging the last link of robot, with the probe moving into the end region directly. The definition of the start region and the end region is based on the semantic information of corresponding ultrasound images. The kidney segments of start-region images are arounding, while those of end-region images are centering. Demonstrations would finish as while as the ultrasound images of standard scanning planes were acquired. In total, 100 sets of data about ultrasound scanning tasks were collected, with 21221 groups of $<X_i,P_i,O_i,F_i>$. Besides, for training the state quality network, the data is labeled by three sonographers, with 4055 groups of acceptable states and 17166 groups of unacceptable states.

\subsection{Pre-training by Imitation Learning}

We employ the deep neural network to learn the ultrasound scanning skills from demonstrated data ($|\theta|=127424519$). The architecture of the neural network is shown in Figure~\ref{fig::network}, and trained parts include front network and action network. The dataset is divide into 8:2 for training and validation, as shown in Figure~\ref{fig::pred_and_true_action}(a). The training process and loss function are presented in \ref{eqn_1} and \ref{eqn_loss}, and the learning rate $\alpha$ is 0.001. The pre-trained network has achieved a good performance in validation set, as shown in Figure~\ref{fig::pred_and_true_action}(b-h). Then, parameters of front network were fixed, and the state quality network was trained to yield confidence of tasks. The learning rate was 0.001 and the loss was calculated by cross-entropy function. Finally, the state quality network achieve the accuracy rate of 98.52\% in training, and 95.04\% in validation. 

\subsection{Human Guided Explorations}

\begin{figure*}[htp]
\centering
\begin{minipage}{1\linewidth}
\centering
\begin{overpic}[width=1\linewidth]{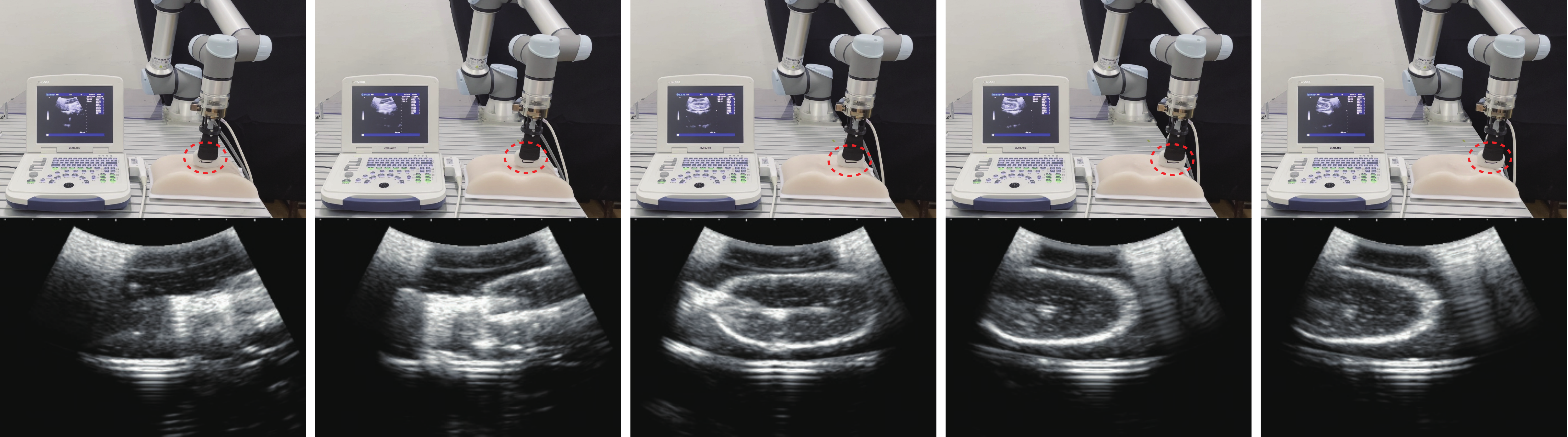}
\put(2, -2.5){\color{black}{(1) Close to target}}
\put(22, -2.5){\color{black}{(2) Close to target}}
\put(42, -2.5){\color{black}{(3) Close to target}}
\put(64, -2.5){\color{black}{(4) Overshoot}}
\put(84, -2.5){\color{black}{(5) Overshoot}}
\end{overpic}\vspace{0.5cm}
\end{minipage}
\caption{\footnotesize Snapshots of the robotic ultrasound scanning tasks guided by the pre-trained model. Here presents the overshoot deficiency of the pre-trained model.}
\label{fig::real_sence_of_model1}
\end{figure*}

\begin{figure*}[hbp]
\centering
\begin{minipage}{1\linewidth}
\centering
\begin{overpic}[width=1\linewidth]{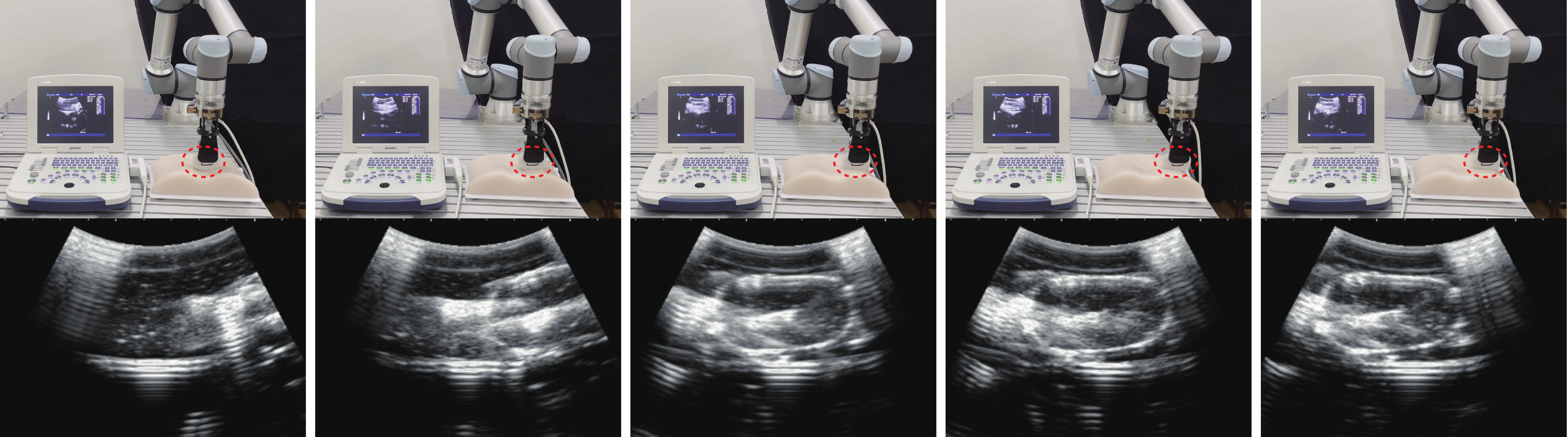}
\put(2, -2.5){\color{black}{(1) Close to target}}
\put(22, -2.5){\color{black}{(2) Close to target}}
\put(42, -2.5){\color{black}{(3) Close to target}}
\put(62, -2.5){\color{black}{(4) Close to target}}
\put(82, -2.5){\color{black}{(5) Stop at target}}
\end{overpic}\vspace{0.5cm}
\end{minipage}
\caption{\footnotesize Snapshots of the robotic ultrasound scanning tasks guided by the optimized model. After optimization, the model basically got rid of the deficiency of overshoot.}
\label{fig::real_sence_of_model2}
\end{figure*}

\begin{figure*}[htbp]
\centering
\includegraphics[width=0.9\linewidth]{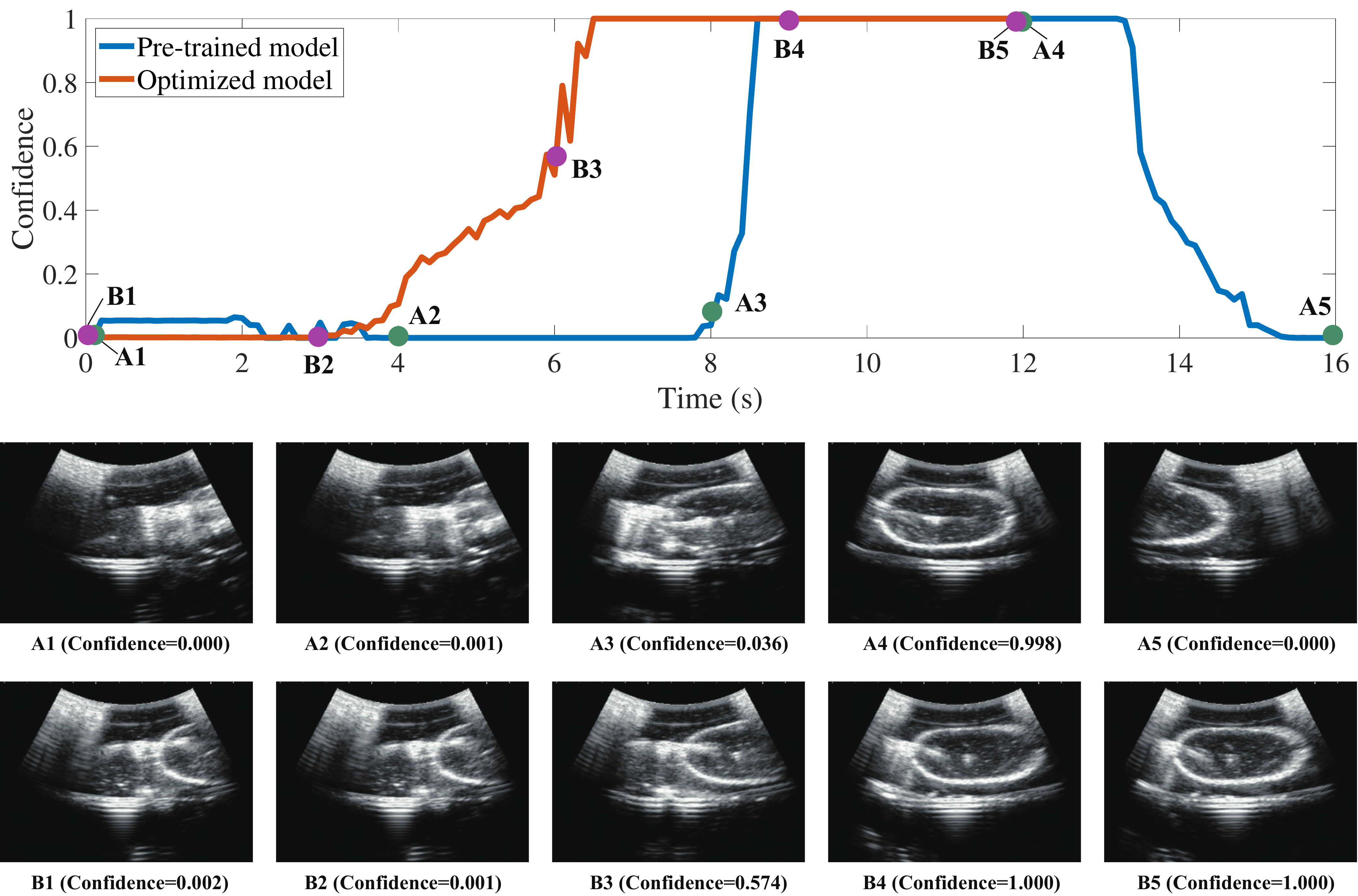}
\caption{\footnotesize The confidence of ultrasound scanning tasks guided by pre-trained model and optimized model. The confidence is the probability of label 1 predicetd by the state quality network, which is defined as the state quality.}
\label{fig::confidence}
\end{figure*}

Though the pre-trained model could preform complete tasks with a naive strategy, its decision-making ability had many flaws, such as overshoot tendency.  Therefore, we optimized the pre-trained model by the human guided explorations, as shown in \ref{eqn_2}. The loss function is shown in \ref{eqn_loss} and the learning rate $\alpha$ is 0.001. The reward function $f_r$ is presented in \ref{eqn_3}, with the trained state quality network for task evaluation. During guided explorations, the pre-trained model would preform autonomous ultrasound scanning tasks under human supervision. For some low-reward actions, the pre-trained would be demonstrated by human suggented actions for optimization. After 200 epochs optimization, the strategy of model had been significantly improved, as shown in Figure~\ref{fig::real_sence_of_model1} and Figure~\ref{fig::real_sence_of_model2}. The confidence of tasks is recorded and shown in Figure~\ref{fig::confidence}, which was defined as the probability of acceptable state predicted by the state quality network. For the pre-trained model, the confidence droped to 0 after reaching a high level, which was caused by overshoot motion. The pre-trained model mastered the correct direction of motion, but lacked the ability to perceive and locate the target. This deficiency has been significantly improved in the optimized model, and its action would converge to a static situation after obtaining a high-confidence state.

\section{Discussion and Conclusion}
\label{sec::discussion_and_conclusion}

\subsection{Discussion}
\label{sec:discussion}
The proposed framework in this paper has the following merits: (1) The ultrasound scanning skill is abstracted as a multi-modal model, which takes ultrasound images, the pose/position of probe, contact force into account. We have also verified that this model has sufficient features for the realization of autonomous ultrasound system. For different robotic ultrasound systems and different target organs, this model can be considered as a reference. (2) The model is pre-trained by imitation learning, and improved by the post-optimization method. For the ultrasound scanning tasks, the model could rapidly learn the demonstrated skills, further be improved by guided explorations. Therefore, this method can present great robustness in our experiment.

Although we have made some achievements in this research, there are still some problems to be solved. First, since the experimental object is the phantom, we have not taken the human interference factors into account, such as breathing and unconscious movement. Second, the robustness and portability of our model need more experiments to prove. Third, the optimized model has not fully grasped the control strategy of the probe pose, which is affected by the probe position and contact force. We have not verified whether this tendency can be improved with more targeted demonstrations and guidance. Our future work will focus on solving the above problems.

\subsection{Conclusion}
\label{sec:conclusion}
This paper presents a two-stage framework to fulfill autonomous ultrasound system. First, we encode the ultrasound scanning skills into a multi-modal learnable model. Further, we propose to use imitation learning for rapid learning of demonstrated skills. Finally, we proposed a post-optimization method to improve the pre-trained model by guided explorations. The experimental results show that the proposed framework can quickly achieve results in training and robustly perform in real environment.

\section{Acknowledgment}
This work was supported by the Natural Science Foundation of Jiangsu Province (Grant No. BK20180235) and the Research Grants Council of the Hong Kong Special Administrative Region, China (Ref. No. 24209021). The experiments have been carried out with help of ultrasound physician Wen Cheng from the Hospital of Wuhan University.

\addtolength{\textheight}{-2cm}   




\bibliographystyle{ieeetr}	
\bibliography{refer} 

\end{document}